# A Stepwise Questioning Expert-Editor Multi-Agent Framework for Long-Document Summarization

Lingyun Shen[1[0000-0001-9054-6493]] and Xuejia Guo[1[0009-0006-6915-3963]]

[1] China South Industry Academy BJ 102209, China
sheryshen20@gmail.com

**Abstract.** Although large language models (LLMs) have shown promising potential in news summarization tasks, their performance on long-document summarization remains challenging as their length often exceeds the input limits. As the agent investment, which provide possibility to improve the inherent capabilities of LLMs. To enhance the effectiveness of long-document summarization based on LLMs, this paper proposes an expert–editor stepwise questioning multi-agent method, in which the expert and the editor guide another agent to refine the summary by posing questions on different aspects of the content and providing targeted clues for revision. We conducted experiments on two representative long-document scientific datasets and evaluated the results through widely recognized automatic metrics. The results demonstrated the effectiveness of our method.

## 1 Introduction

Large language models (LLMs) have achieved remarkable success across nature language understanding and generation task.(Sun et al., 2021; Chowdhery et al., 2023; Brown et al., 2020) Leveraging LLMs for summarization with Instruction has shown significant potential(Zhang et al., 2023; Liu et al., 2025; Li et al., 2025a). However, owing to the context processing capability and hallucination issues, LLMs ability on long document summarization remains need further improved(Wan et al., ; Chan et al., 2023; Liu, 2024). Recently, methods that leverage multi-agent systems to enhance the intrinsic abilities of large language models have shown promising potential in various long-form generation and reasoning tasks(Gu, 2025; Wan et al., ; Hu, 2025; Fang et al., ; Mo and Hu, 2024). This suggests that multi-agent collaboration and planning may address the limitations in long-document summarization based on LLMs.

Currently, various agent framework are work on leading LLMs particpatly subtask by prompt technologies(Mo and Hu, 2024; Fang et al., ; Wan et al., ; Hu, 2025). Prior prompt technologies studies provides lots of experience that utilize large language models to construct agents, such as role play(Schmidt et al., 2023; Wang et al., 2024c; Zheng et al., 2024)， decompose a complex problem into a series of simple

questions(Press, 2023), iterative refined(Jha et al., 2023; Bakharia, 2025),etc. Based on empirical observation, we consider that stepwise questioning may stimulate LLMs to revie their previously generated output. As shown in figure1, the question “What are the key terminologies in the original article?” triggered the large language model to reflect on and revise its prior summary.

Inspired by these prior studies and practical experiences, we propose a stepwise questioning multi-agent approach for long-document summarization. By capturing key information more accurately to grasp the core idea of the article (Shen and Le, 2023), this method can address the issues of incomplete information and unfaithful in long document summarization base on LLMs. Specifically, We design two primary roles, expert who foucuses on semantic understanding and the important clue of the main idea about article for generating consistent, fluently and faithful summary, and editor who focuses on the surface-level literal errors such as grammatical and syntactic accuracy, as well as consistency with the original text. They both guide the author agent to reflect on and revise its initial summary written previously through progressively asking questions. Both guide another Author agent through step-by-step questioning to reflect on and revise the initially written summary. This idea is inspired by the academic review process,where experts pay attention to the research contents of text and editors care more about the format and other surface-level errors to ensure the reliability of the entire text. We conducted experiments on two common long document datasets,Arxiv and pubmed(Cohan et al., 2018).The result demonstrate that our method can effectively improve long document summarization compared to traditional segmenting-and-aggregating approaches. And multi-agent interaction also reduce the generation of irrelevant content.

The main contributions of this paper can be summarized as follows:

- We propose an Expert-Editor stepwise questioning multi-agent method for long document summarization. Inspired by academic review processes and ask LLMs iteratively methods, which stimulates LLMs to reflect and optimize generated summaries.
- We designed a list of crucial questions from aspects of the quality and consistency of abstract summary. Then we analyzed the effectiveness of these questions in improving summarization.
- We conducted experiments on advanced commercial and open source models respectively. Experimental results demonstrate that our method superior prior approaches for long document summarization.

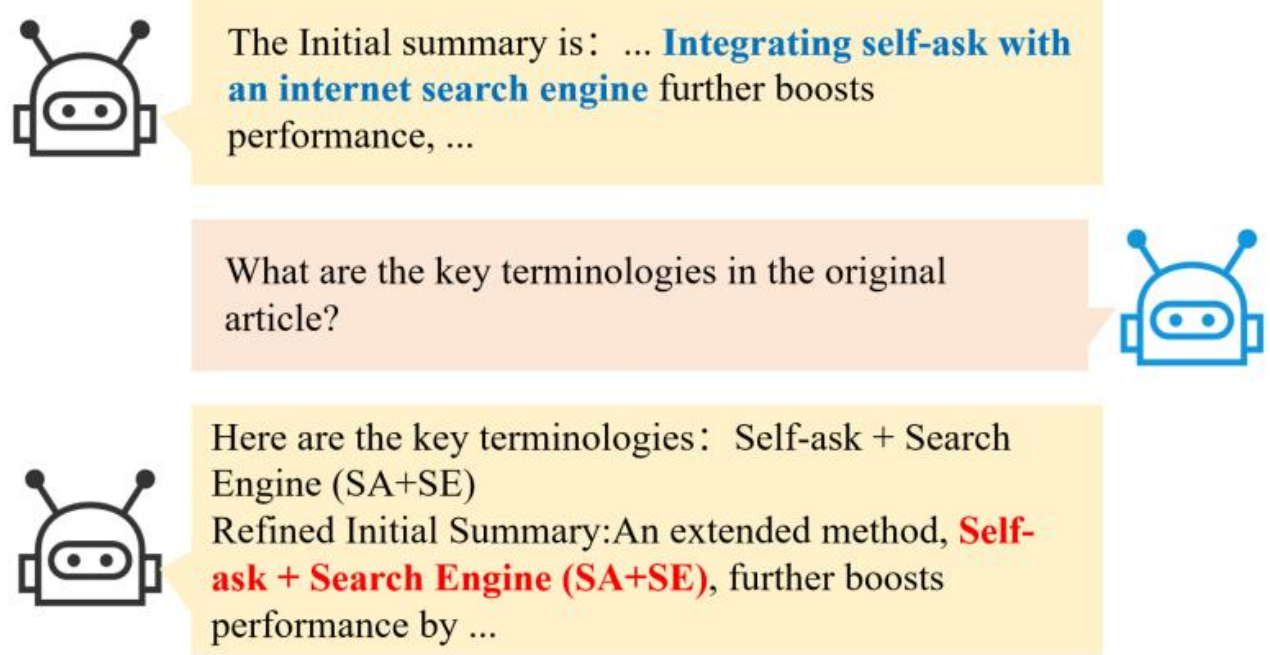

**Fig. 1.** A case of stimulating refinement by asking question.

# 2 Related works

## 2.1 Long Document Summarization with Large Language Models

The main challenges of long document summarization are attributed to the excessive length of the context, which leads to information loss and related issues(Dong, 2024; Hsieh et al., 2024). Previous research trained pre-trained models based on Transformer architectures increase limited due to low quality data . (34Efficient Attentions for Long Document Summarization; 31 Factorizing Content and Budget Decisions in Abstractive Summarization of Long Documents).Previous studies have primarily addressed the inability to input long texts into LLMs in a single pass by employing the segment-and-aggregate approach. Li et al., (2025) segmented long documents into paragraph-level text chunks, iteratively querying clues about main events from paragraphs to construct the primary content of articles. (Cho et al., 2022) divided articles by sections, preserving structural information and determining main content by selecting important sentences from each section. These method for improving summary quality remains limited.

With the success of large language models, the training cost for models has been reduced(Wang et al., 2024b). LLMs motivated more low-resource solutions for text summarization(Sahu et al., 2025a; Zhu et al., ). (Sahu et al., 2025b)) employed large models to extract key information from documents for summary generation under 50-shot settings, outperforming BERT-based pre-trained model. Zhang et al., (2023) compared the effectiveness of different LLMs for text summarization generation, revealed that instruction models could generate higher-quality summaries that better align with human preferences.

## 2.2 Multi-Agent Systems Using Prompting Techniques

Large language models show capabilities that compete with human intelligence in language understanding and generation(OpenAI et al., 2024; Touvron et al., 2023b; Touvron et al., 2023a), thus providing support for inter-LLM interaction and autonomous decision-making.. Nascimento et al., (2023) proposed that GPT(Brown et al., 2020; OpenAI et al., 2024) can enhance individual models' intrinsic capabilities through multiple rounds of iteration and synergy, thereby result in refined generated content.Multi-agent systems utilize interactive frameworks composed of multiple models to achieve complex tasks, typically which driven by carefully designed prompts (Schulhoff et al., 2025). Another important characteristic of multi-agent systems is role-playing (Wang et al., 2024a), where different roles and personas are assigned to each agent, instructing models to complete different subtasks. This kind of system construction approach performance remarkable in many complex and long-form tasks like long-text question answering(Wan et al., 2025), text simplification(Fang et al., ), and complex reasoning(Gu, 2025).Wan et al., (2025) constructed a Detective-Critique-Refine multi-agent method that optimizes output results for different subtasks through multi-round debates, demonstrating the effort of multi-agent systems on long-text tasks.

However, research investment addressing information loss and consistency issues in long document summarization based on multi-agent remains insufficiently studied.

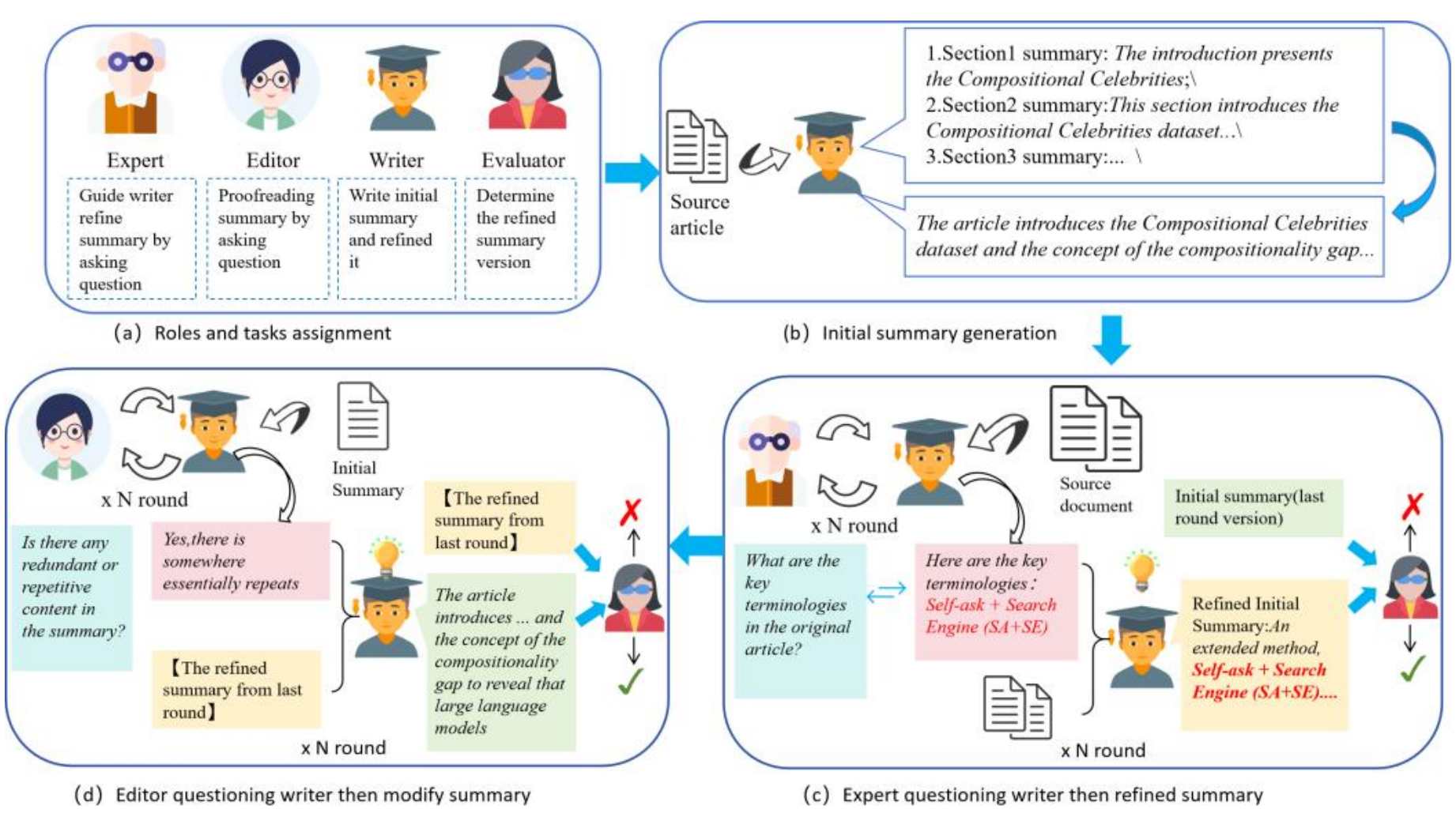

**Fig. 2.** The overview of expert-editor stepwise questioning multi-agent.

## 3 Methodology: An Expert-Editor Stepwise Questioning Multi-Agent

We simulate the process of students reading understanding in exam and flow of academic review by asking questions to guide writer agent identify the essential information relevant to the summary. After answer the expert's questions, student writer both obtains the salient clue of the main idea of article and gains inspiration for writing the summary. This process can take the place of reading the entire text. We design questions to be as simple as possible to minimize additional errors that may occur when LLMs answer complex questions(Press, 2023). Our multi-agent overall workflow is illustrated in Figure 1, which includes four steps:(1)Roles and tasks assignment （2）Writer agent write the initial summary. (3)Experts and editor formulate the questioning plan.(4)Refinement of the initial summary through expert and editors asking question to writer agent. The way of collaboration among agents follows a pipeline communication strategy. Each step of whole process described in detail below.

### 3.1 Roles and Tasks Assignment

To effectively achieve our objective, we design four roles for the method, they are expert, editor,writer and evaluator. The expert focuses on the semantic content of the summary while the editor attends to proofreading and the correction of linguistic errors to enhance overall quality. They formulate each questioning plan accordingly, with one oriented toward the main ideas of the content and the other toward the linguistic aspect.The author drafts the initial summary and responds to the expert and editor's questions then refines his previously written summary. After each revision, the evaluator assesses the updated version against the previous one, determining whether the changes improve quality.

### 3.2 Initial Summary Generation

We adopted a straightforward approach by feeding articles directly into large language models (LLMs) and instructing them to generate summaries.Next,We follow the methods in previous studies(Li et al., 2025b; Zhong and Litman, 2025), divide the whole document into a series of text chunks by sections, let the writer draft a local summary for each section, and then aggregate an overall summary for the source document according to the local summaries. This overall summary serves as the initial version for the next expert questioning stage.

### 3.3 Questioning Plan Formulation

Before initiating the questioning and refinement process, experts and editors formulate a questioning plan based on the article and its initial summary. We introduce a set of questions designed around empirically intuitive, qualitative evaluation criteria for summaries—completeness, coherence, relevance, and factual consistency (Zhang et al., 2024; Zhong & Litman, 2025). Notably, not every question is applied in refining a given summary. The expert and editor agents determine the list of questions and their sequence based on the specific context of each summary.

### 3.4 Summary Refinement through Expert and Editor Questioning

To effectively achieve our objective, we designed four core roles—expert, editor, author, and evaluator—for the proposed method, where the expert focuses on the summary's semantic content and the Editor is responsible for proofreading inconsistencies with the original text and correcting linguistic errors to enhance overall quality and consistency. Both roles pose questions incrementally based on the query plan formulated in previous steps. The author responds to questions raised by the expert or Editor, and after each response, reflects on whether revisions to the previous summary are necessary. The answers to the expert's questions convey the key information implicit in the source article, which are useful to revise the previous version of the summary. Subsequently, the evaluator selects the better summary between the previous version and the revised one using criteria of completeness, coherence, relevance, and consistency. After multiple rounds of question-answering-reflecting-revise, the final output as the refined summary in this stage.

Next, the editor proceeds to ask questions.To avoid unnecessary revisions, the editor's questions adopt a yes-no format.The author answers these questions by compare the summary with the source article. If the issue in the editor's question exists in the current summary, the author responds "yes" and revises the summary. Otherwise,if the issue does not exist, the author responds "no" without any modifications. The Evaluator then compares the two versions (previous and revised) and selects the better one based on the same criteria. After multiple rounds of such QA interactions, the final revised summary is deemed the refined result.

# 4 Experiments

## 4.1 Datasets

Our experiments were conducted on two representative long-document datasets—arXiv and PubMed (Cohan et al., 2018).These datasets consist of original scientific texts and include the structural information of the articles. Scientific texts are considerably longer than those in news datasets. We computed the length statistics of the datasets, rounded them to a fixed number of decimal places, and presented the results in Table 1. For the automatic evaluation, we randomly sampled 300 instances as test set.

**Table 1.** Length statistics of datasets.

| Dataset | Article | | Summary | |
|---|---|---|---|---|
| | Avg. Words | Avg. Sents | Avg. Words | Avg. Sents |
| Arxiv | 5171.85 | 206.3 | 257.94 | 9.61 |
| Pubmed | 2872.13 | 206.3 | 194.22 | 6.85 |

## 4.2 Models

Our agent-based method is built upon large language models, and in our experiments we conducted studies using both commercial and open-source models. Specifically,we employed, DeepSeek-R1-0528[1], LLaMA 3.1 8B and LLaMA 3.1 70B[2].

## 4.3 Evaluation Metrics

We employed three automated metrics to evaluate the relevance, completeness, and factual consistency of the generated summaries: ROUGE-L (Lin, 2004), BERTScore(Zhang et al., 2020), and FactCC (Kryściński et al., 2019). ROUGE-L evaluation was conducted using ROUGE-1, ROUGE-2, and ROUGE-L scores.

## 4.4 Experimental Implementation Details

To prevent infinite questioning and revisions, we conducted preliminary experiments on a small test set and found that effective questions—i.e., those questions leading to better summaries after refinement—are not infinite. Therefore, our experimental setup limits the expert to a maximum of 5 questions per round and the editor to 4 questions. To reduce contextual bias for the evaluator agent, we randomly shuffled the order of the original and revised summaries before presenting them for evaluation.

We used NVIDIA A100 80GB GPUs as the computational resources for our experiments on LLaMA 3.1 8B and LLaMA 3.1 70B.

---

[1] https://www.deepseek.com/
[2] https://www.llama.com/models/llama-3/

**Table 3.** Main Results of Expert-Editor Stepwise Questioning Multi-Agent.

| Model | Dataset | Method | R-1 | R-2 | R-L | BS | FC |
|---|---|---|---|---|---|---|---|
| LLama-3.1-8B | Axiv | DG | 32.22 | **8.23** | 16.25 | 80.41 | 54.33 |
| | | HERA(Li et al., 2025a) | 30.03 | 7.51 | 15.39 | 80.28 | **70.00** |
| | | SQ2E | **34.19** | 8.07 | **16.89** | **81.38** | 60.67 |
| LLama-3.1-70B | | DG | 33.53 | 9.44 | 17.65 | 80.99 | 61.00 |
| | | HERA(Li et al., 2025a) | 37.23 | 10.66 | **19.06** | 82.19 | 47.33 |
| | | SQ2E | **38.89** | **11.27** | 18.95 | **84.01** | 50.67 |
| Deepseek-R1 | | DG | 36.19 | 10.64 | 17.88 | 81.28 | 46 |
| | | HERA(Li et al., 2025a) | 36.78 | 10.28 | 19.64 | **82.49** | **57** |
| | | SQ2E | **38.94** | **11.74** | **20.22** | 82.33 | 52.67 |
| LLama-3.1-8B | Pubmed | DG | 32.22 | 8.23 | 16.25 | 80.41 | 54.33 |
| | | HERA(Li et al., 2025a) | 29.66 | 8.11 | 16.26 | 81.90 | 59.33 |
| | | SQ2E | **34.65** | **9.16** | **17.35** | **82.41** | 55 |
| LLama-3.1-70B | | DG | **39.58** | **14.25** | **20.7** | 82.86 | **58.33** |
| | | HERA(Li et al., 2025a) | 35.34 | 11.31 | 19.67 | 83.14 | 37.33 |
| | | SQ2E | 36.82 | 12.28 | 20.60 | **83.87** | 54 |
| Deepseek-R1 | | DG | 35.43 | 11.03 | 17.70 | 81.82 | 34.33 |
| | | HERA(Li et al., 2025a) | 31.03 | 9.57 | 18.04 | **83.27** | **38.67** |
| | | SQ2E | **37.29** | **12.04** | **19.86** | 83.21 | 35.67 |

*Note that Automatic metrics BS, FC, R-1, R-2, and R-L correspond to BERTScore, FactCC, ROUGE-1, ROUGE-2, and ROUGE-L, respectively. Compared methods: DG (Directly Generation), HERA (Li et al., 2025a), and our proposed SQ2E Multi-Agent.

### 4.5 Main Results

The main results are presented in Table 2, with the highest scores highlighted in bold. Overall, our proposed method is effective in improving the accuracy and factual consistency of summary. In particular, when using LLama-3.1-8B and Deepseek-R1 models, significant improvements are observed in ROUGE, BERTScore, and FactCC metrics. However, the scores of some metrics decrease noticeably on LLama-3.1-70B. We conjecture that this phenomenon may be attributed to the loss of precision during the inference of the 70B-parameter model, which is associated with the computational resources utilized in our experiments. Thus, the experimental results demonstrate that the method proposed in this paper can enhance the summarization ability of large language models (LLMs).

### 4.6 Results Analysis

To explore the effectiveness of the experiment, we counted the number of valid questions used by the model to refine summaries for each dataset. A valid question is defined as one where the expert or editor raise a question, the author revised the summary accordingly, and the evaluator judged the revised summary superior. This result is presented in Figure 3. As shown in the figure, the questions raised by the expert and editor played a significant role in optimizing the model-generated summaries. In most cases, the expert's questions led to 3-4 valid optimizations, and the editor's questions resulted in 2-3 valid revisions. This confirms the effectiveness of the question list designed in this study. From the perspective of automatic metrics, the final results are not positively correlated with model parameters. This is partly due to computational resource constraints and partly because small-parameter large language models (LLMs) can achieve performance comparable to that of large-parameter models through well-designed prompt engineering.

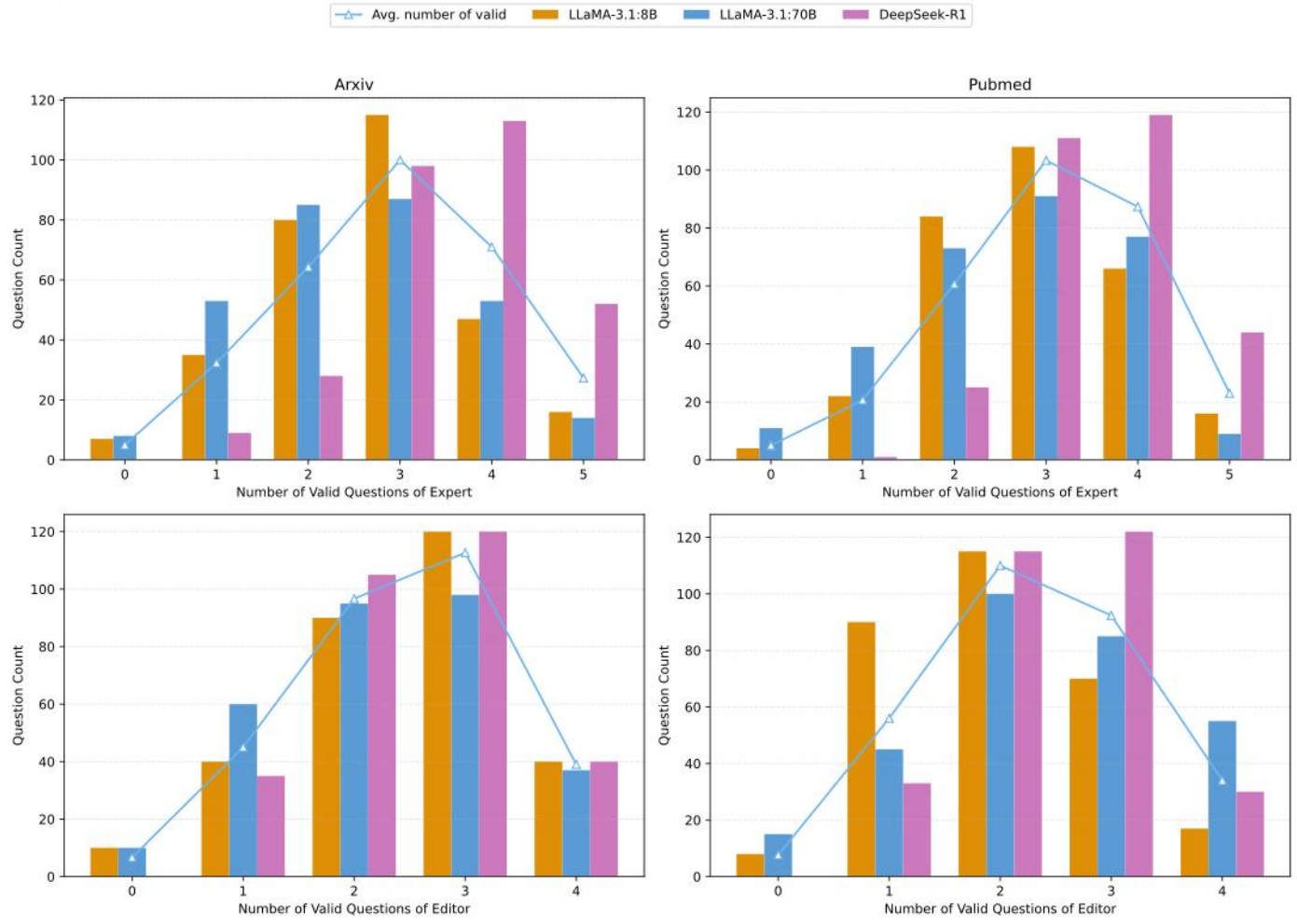

**Fig. 3.** The number of effective questions used by experts and editors to refine summaries, evaluated with three models on the ArXiv and PubMed datasets. The open triangles denote the average count of effective questions.

## 5 Conclusion

This paper propose an expert-editor stepwise questioning multi-agent for long-document summarization. The author agent, progressively questioned by the expert and editor, was prompted to obtain clues to key information for the summary and to refine the initial version in a targeted manner. To construct an effective questioning plan, A well-crafted question bank was designed for the Expert and Editor to guide the formulation of their query lists. The experimental results demonstrate that summary refinement through questioning can improve simple summary generation via segmentation and aggregation.